\documentclass[10pt,twocolumn,letterpaper]{article}

\usepackage[pagenumbers]{cvpr} 

\usepackage{graphicx}
\usepackage{amsmath}
\usepackage{amssymb}
\usepackage{booktabs}
\usepackage{diagbox}

\usepackage[pagebackref,breaklinks,colorlinks]{hyperref}

\usepackage[capitalize]{cleveref}
\crefname{section}{Sec.}{Secs.}
\Crefname{section}{Section}{Sections}
\Crefname{table}{Table}{Tables}
\crefname{table}{Tab.}{Tabs.}

\begin{document}

\title{Unsigned Distance Field as an Accurate 3D Scene Representation \\ for Neural Scene Completion}

\author{
Jean Pierre Richa$^{1,2}$, Jean-Emmanuel Deschaud$^{1}$, François Goulette$^{1}$ and Nicolas Dalmasso$^{2}$ \\
\textit{$^{1}$ MINES ParisTech, PSL University, Centre for Robotics, 75006 Paris, France} \\
\textit{$^{2}$ ANSYS France, 15 Pl. Georges Pompidou, 78180 Montigny-le-Bretonneux, France} \\
}

\newcommand{\JP}[1]{{ \color[rgb]{1,0,0} {{\sc JP said:} #1}}}

\maketitle

\begin{abstract}
    Scene Completion is the task of completing missing geometry from a partial scan of a scene. Most previous methods compute an implicit representation from range data using a Truncated Signed Distance Function (T-SDF) computed on a 3D grid as input to neural networks. The truncation decreases but does not remove the border errors introduced by the sign of SDF for open surfaces. As an alternative, we present an Unsigned Distance Function (UDF) as an input representation to scene completion neural networks. The proposed UDF is simple, and efficient as a geometry representation, and can be computed on any point cloud. In contrast to usual Signed Distance Functions, our UDF does not require normal computation. To obtain the explicit geometry, we present a method for extracting a point cloud from discretized UDF values on a sparse grid. We compare different SDFs and UDFs for the scene completion task on indoor and outdoor point clouds collected using RGB-D and LiDAR sensors and show improved completion using the proposed UDF function.
\end{abstract}


\section{Introduction} \label{sec: intro}
    
    Distance functions (DFs) computed on regular grids provide rich information about the geometry of a scene. Many DFs have been proposed to tackle a wide variety of tasks, such as surface reconstruction~\cite{Hoppe1992, oztireli2009}, Simultaneous Localisation And Mapping (SLAM)~\cite{Newcombe2011KinectFusion, Deschaud2018IMLSSLAM}, path planning~\cite{Oleynikova2017Voxblox} and scene completion~\cite{sscnet, sgnn}. Traditional surface reconstruction methods~\cite{Hoppe1992, Curless1996VRIP, imls} compute a Signed Distance Function (SDF) representation from range images or point clouds on a regular volumetric grid. A zero level set extraction algorithm such as Marching Cubes~\cite{marching-cubes} can be deployed to extract a final mesh of the scene. SDFs fill small holes by local interpolation, however, they fail to fill large missing areas, such as occluded regions caused by dynamic objects in the foreground, or holes caused by the sparsity of points.
    
    \begin{figure}[t]
        \centering
        \includegraphics[width=\linewidth]{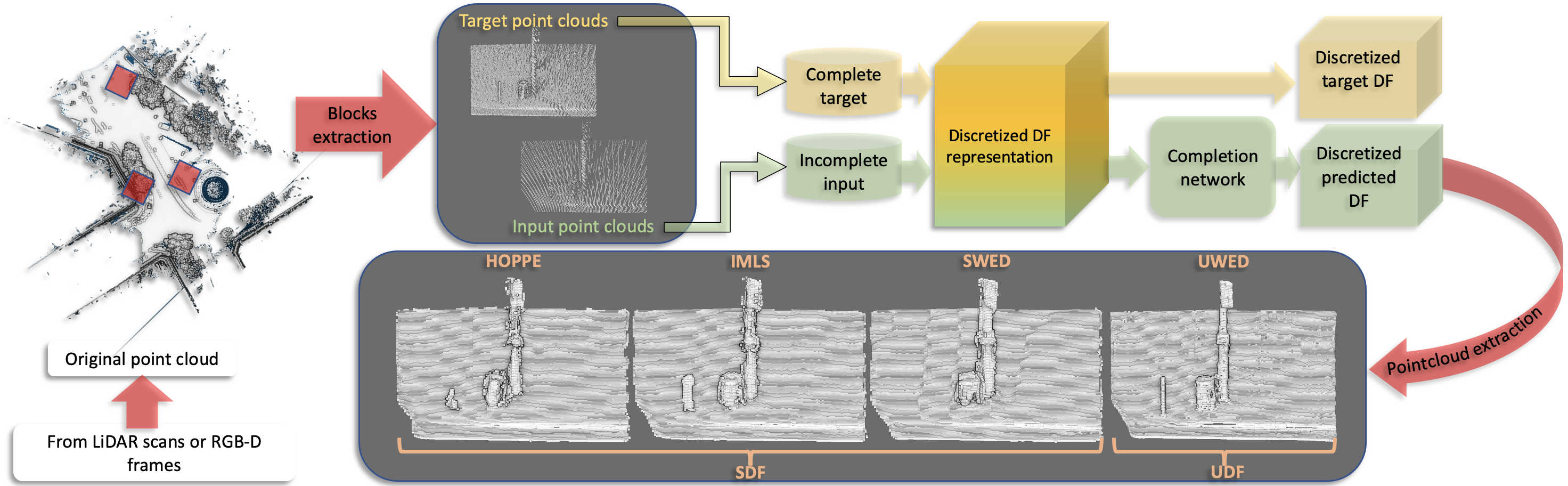}
        \caption{Self-supervised 3D scene completion comparing several Signed Distance Functions (SDFs) and our Unsigned Distance Function (UDF) called UWED for Unsigned Weighted Euclidean Distance. Our UDF eliminates the sign ambiguities present in SDFs achieving higher accuracy on the scene completion task.}
        \label{fig:scene_completion_pipeline}
    \end{figure}
    
    With advances in learning-based methods, recent studies have looked to develop approaches tackling the problem of larger missing areas, which gave rise to the scene completion task. Different data representations have been proposed that focus primarily on 3D binary occupancy grids, point clouds, or SDFs representing the underlying surface. Occupancy voxel grids do not allow the geometry to be finely represented~\cite{3D-EPN, scancomplete}. Point cloud completion networks, such as PCN~\cite{PCN}, or more recently, VRC-Net~\cite{pan2021}, are limited to single objects and are not able to deal with large scenes. In contrast, using Truncated Signed Distance Function (T-SDF) as a surface representation has proven to be able to accurately represent and predict the geometry of large scenes in a variety of previous works~\cite{sscnet, VVNet, Chen2019Adversarial, Zhang2019Completion, Chen2019Attention, scancomplete, sgnn, Dai_2021_CVPR}.
        
\begin{figure}[ht]
    \centering
        \includegraphics[width=\linewidth]{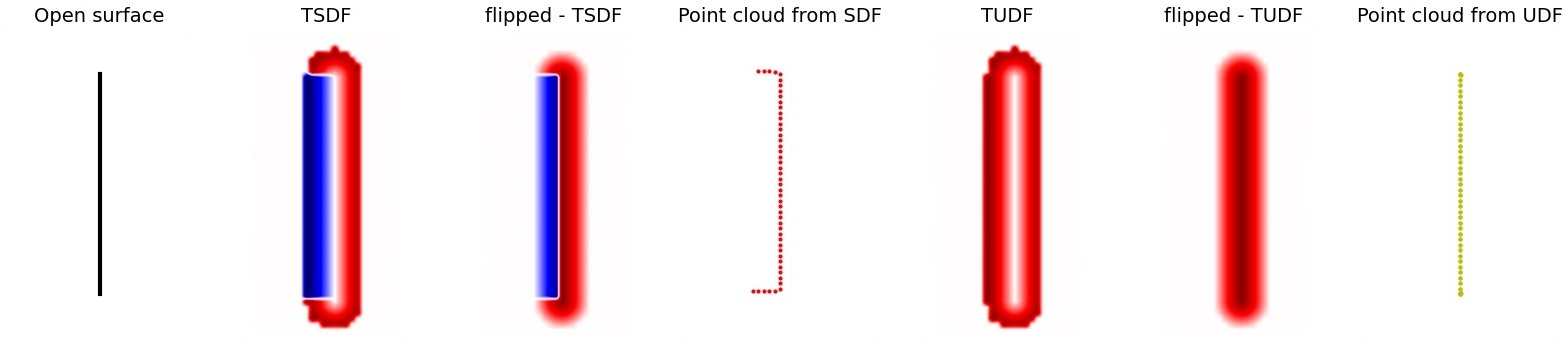}
        \caption{A 2D simple example to show the interest of using our UDF. Even if the truncation limits the effects, the extraction of a point cloud from a T-SDF generates errors on borders (point cloud in the middle). An Unsigned Distance Function (UDF) eliminates the problem of negative and positive sides (point cloud on the right). 
        }
        \label{fig:TUDF_2D}
    \end{figure}

    One issue in using SDFs is their need to perform inside/outside classification of the volume for open shapes. For point clouds, such as those obtained from real-world scenes, this issue can be solved by computing point-wise oriented normals (a challenging task when done on noisy point clouds~\cite{schertler2017, metzer2021}) and truncating the SDF values resulting in a Truncated SDF that limits the interference between spatially nearby surfaces. However, T-SDF does not eliminate the geometry dilation on borders (3D example in Figure~\ref{fig:scene_completion_pipeline} and 2D example in Figure~\ref{fig:TUDF_2D}). 
        
    To address these issues, we propose the computation of an Unsigned Distance Function (UDF), namely, Unsigned Weighted Euclidean Distance (UWED) on a 3D volumetric grid directly from point clouds, which can then be applied on scenes acquired using RGB-D or LiDAR sensors. However, the extraction of an explicit geometry such as a mesh from a UDF is not resolved~\cite{Congote2010MarchingCI}. Thus, we present a method for extracting point clouds from unsigned distance fields. We show that the proposed unsigned distance representation allows scene completion networks to better understand the geometry of the scene, without having to manage the sign, and predict a better completion of the scene.
    
    We perform extensive tests of signed and unsigned DFs with two different state-of-the-art scene completion networks: SG-NN~\cite{sgnn} able to predict new SDF/UDF values from which we can extract a point cloud, and another 3D U-Net using Minkowski generalized sparse convolutions~\cite{minkowski} (MinCompNet for MinkowskiCompletionNet) able to predict a binary occupancy grid.
        
    Our contributions are summarized as follows:
    \begin{itemize}
        \item A simple, and robust unsigned distance function, UWED, as input for neural scene completion, and a method to extract a point cloud from an unsigned distance field computed on a sparse 3D grid.
        \item Extensive comparisons of scene completion using several signed and unsigned distance functions on various environments (indoor and outdoor), with different types of datasets (real and synthetic), acquired using different sensors (RGB-D and LiDAR), and validated on two different network architectures, namely SG-NN~\cite{sgnn} and MinCompNet~\cite{minkowski}.
    \end{itemize}

\section{Related Work}
    
    \paragraph{\textbf{Signed Distance Functions}}
    Using DFs on 3D volumetric grids for surface reconstruction traces back to the seminal work of Hoppe \textit{et al.}~\cite{Hoppe1992}, in which a Signed Distance Field $\phi: \mathbb{R}^3 \rightarrow \mathbb{R}$ was used to represent the underlying surface in a point cloud:
    \begin{equation} \label{eq: SDF}
        \phi(x) = \mathbf{n}_i \cdot (x - \mathbf{p}_i)
    \end{equation}
    where $x$ is the voxel coordinates in the 3D grid, $\mathbf{p}_i$ is the nearest neighbor of $\mathbf{x}$ in the point cloud $P$, and $\mathbf{n}_i$ is the normal unit vector associated with the point $\mathbf{p}_i$. 
    
    Having computed the SDF on a regular 3D grid, the Marching Cubes algorithm~\cite{marching-cubes} extracts the final iso-surface as a mesh. 
    
    Following this pioneering work, several methods were proposed to deal with noisy data, such as Implicit Moving Least Squares (IMLS)~\cite{imls}, which approximates the local neighborhood of a given voxel in the grid as a weighted average of the local point functions:
    \begin{equation}\label{eq: imls}
        \text{IMLS}(x) = \frac{ \sum_{\mathbf{p}_{k} \in N_{x}} \mathbf{n}_k \cdot (x - \mathbf{p}_{k})   \hskip.03in \theta_{k}(x) }{\sum_{\mathbf{p}_{k} \in N_{x}} \theta_{k}(x)}
    \end{equation}
    where $x$ is the voxel coordinates, $N_{x}$ is the set of $\mathbf{p}_k$ neighboring points from $x$ in the point cloud $P$, $\mathbf{n}_k$ is the normal unit vector associated with the point $\mathbf{p}_k$, and $\theta_{k}$ is the Gaussian weight defined as follows:
    \begin{equation}\label{eq: gaussian weight}
        \theta_{k}(x) = e^{-||x - \mathbf{p}_k||_{2}^{2} / \sigma^{2}}
    \end{equation}
    where $\sigma$ is a parameter of the influence of points in $N_{x}$. 
    
    In a different approach, the volumetric fusion method of VRIP~\cite{Curless1996VRIP} takes advantage of range images by meshing the depth image to cast a ray from the sensor origin to the voxel of the volumetric grid, obtaining a signed distance to the mesh, and then merging the scans distances in a least-squares sense. However, this DF can only be computed from range images and cannot be used directly on point clouds.
    
    Other surface reconstruction methods~\cite{Kazhdan2006Poisson, Kazhdan2013poisson} use global implicit functions such as indicator functions where the reconstruction problem is solved using a Poisson system equation. Finally, some methods use 3D Delaunay tetrahedralization~\cite{labatut2009, zhou2019, caraffa2021}, and more recently, others apply 3D ConvNets~\cite{peng2020, ummenhofer2021}.
    
    All these "surface reconstruction" methods mainly change the geometry of the scene from a raw representation (depth images or raw point clouds) to a more usable geometry such as meshes. They attempt to produce watertight meshes, however, they do not try to fill large missing regions like scene completion networks. 
    
    \paragraph{\textbf{Unsigned Distance Functions}}
        
        UDFs solve the ambiguity introduced by SDFs, since they eliminate the need to define an interior and exterior using the sign. However, it is difficult to extract a well-behaved final mesh from such functions. One method~\cite{Congote2010MarchingCI} proposes a variant of the Marching Cubes to be used on an unsigned distance field, but the method results in holes in the reconstructed final mesh. Another method~\cite{signing-the-unsigned} attempts to obtain a consistent signed distance field from unsigned distances on the discretized grid. However, this approach comes with the drawback of a high computational cost, and is not scalable to large scenes.
        SAL~\cite{atzmon2020} trains a fully-connected network to infer the sign of a DF from unsigned data but shows this only on datasets of small objects.
        NDF~\cite{NUDF} and DUDE~\cite{Venkatesh2020DUDE} learn a volumetric unsigned distance encoded in a neural network. 
        The last three methods represent the scene through neural networks, and are known as Implicit Function Learning (IFL) methods.
    
    \paragraph{Implicit Function Learning} 
        One of the first IFL methods, DeepSDF~\cite{deepsdf}, learns a continuous SDF representation from point clouds of carefully scanned closed shapes and unlocks the ability to model multiple classes of objects with the same network. IM-NET~\cite{im_net} designs an implicit decoder that takes the point cloud and a latent vector encoding the underlying shape as input. It predicts an indicator function performing binary, inside/outside classification and then extracts the final iso-surface using the marching cubes algorithm~\cite{marching-cubes}. DMC~\cite{dmc} predicts the final mesh from a point cloud input in an end-to-end fashion instead of predicting the implicit SDF representation and using a surface extraction algorithm. ONeT~\cite{ONet} learns the implicit function and predicts binary occupancy, enabling surface extraction through subdividing the occupied voxels using an octree structure. Then it predicts a finer inside/outside classification, which is later used to extract the final mesh using marching cubes. IF-Nets~\cite{ifnet} learns the implicit function from multi-scale point encoding and predicts a binary classification based on local and global shape features. 
        
        To solve the problems caused by SDFs, NDF~\cite{NUDF} and DUDE~\cite{Venkatesh2020DUDE} learn a volumetric unsigned distance that regresses a UDF from deep features instead of binary occupancy and enables the extraction of more accurate sub-voxel resolution of the explicit surface from the predicted distance field. NDF extracts a dense point cloud by projecting points using the gradient field from the back-propagation through the network, while DUDE also learns a normal vector field along with the DF.
        
        Although IFL methods result in a good final surface representation quality while completing small parts, they do not complete large missing regions caused by foreground or dynamic objects. It is then necessary to use scene completion networks to perform large missing regions completion.
    
    \paragraph{\textbf{Scene completion Networks}}
        The increased availability of range sensors and advances in learning-based methods have made it possible to obtain large amounts of data and train neural networks to predict larger missing areas caused by occlusions from the foreground and dynamic objects. 
        
        Two main approaches to geometry completion exist in the literature. Object-level completion~\cite{PCN, 3D-EPN, PointFlow} focuses on the completion of single objects, such as chairs, cars, and tables. These approaches implicitly learn strong prior, e.g., shape symmetries, about the categories of objects, but they do not scale to large scenes, which limits their use to simple objects. Scene-level completion attempts to complete the missing geometry in large scenes with no strong prior on the objects in the scenes.
    
    \paragraph{\textbf{Scene completion from RGB-D sensors}}
    Scene completion from a single depth image was introduced to complete missing scenes in indoor settings with SSCNet~\cite{sscnet}. This paved the way for completion beyond traditional surface reconstruction methods, which increased interest in the task of scene completion~\cite{ASSCNet, efficient-SSCNet-with-spatial-group-convolution}. These earliest methods focused on completing the scene from a single depth image and used a T-SDF as their scene representation, with T-SDF being the signed distance to the closest point on the surface, similar to the representation of Hoppe \textit{et al.}~\cite{Hoppe1992}. They used a flipped version of the T-SDF to remove discontinuities around the truncation of the function and get strong gradients around the surface, giving more signal to the network around the surface. When multiple range images are available, such as in Matterport3D~\cite{Matterport3D}, most methods use the volumetric fusion VRIP~\cite{Curless1996VRIP} to obtain the discretized T-SDF representation of the scene. They also use different network architectures to infer the missing geometry. For instance, ScanComplete~\cite{scancomplete} uses a U-Net architecture on dense volumes, and SG-NN~\cite{sgnn} uses a U-Net with sparse convolutions. 
    
    Some methods perform semantic scene completion (survey in~\cite{roldao2021}), in which they complete the scene, while predicting per-voxel semantic class, such as ScanComplete~\cite{scancomplete}. Others predict color information, such as SPSG~\cite{Dai_2021_CVPR}, which is an extension of SG-NN~\cite{sgnn}. Our work focuses on improving the geometry representation. However, it could be easily integrated with these works by adding semantic or color information.
    
    \paragraph{\textbf{Scene completion from LiDAR sensors}}
    Fewer methods have been developped for scene completion on LiDAR point clouds. LMSCNet~\cite{lmscnet} proposes a lightweight 2D U-Net backbone that is used on the x, y dimensions to reduce the computation complexity and complete a scene from a single LiDAR scan. This method provides fast inference, however, it is not optimal for accurate geometry prediction since the input is an occupancy grid. Another method~\cite{s3cnet} performs Bird's Eye View (BEV) projection and extracts sparse 2D features along the x-y plane to perform depth completion. They subsequently extract normal information, which is later used to compute the T-SDF values on a 3D regular grid. The inference of the network is also a 3D occupancy grid. However, previous works~\cite{3D-EPN, scancomplete} have shown that the use of signed distance functions (T-SDF) allow better learning of geometry than the use of occupancy grids. 


\section{UDF representation and point cloud extraction}

    In this section, we present the proposed UDF, Unsigned Weighted Euclidean Distance (UWED), that we use as an input for neural scene completion and a method to extract an explicit geometry from UDFs in the form of point clouds.
    
    \subsection{Proposed Unsigned Weighted Euclidean Distance (UWED)} \label{sec: TUDF}
    
        UWED is defined as a weighted average of Euclidean distances and is computed as follows:
        \begin{equation}\label{eq: uwed}
            \text{UWED}(x) = \frac{\sum_{\mathbf{p}_k \in N_{x}} ||x - \mathbf{p}_k||_2 \; \theta_k (x)}{\sum_{\mathbf{p}_k \in {N_{x}}} \theta_k(x)}    
        \end{equation}
        where $x$ is the voxel coordinates, $N_{x}$ is the set of $\mathbf{p}_k$ neighboring points from $x$ in the point cloud $P$, and $\theta_{k}$ is the Gaussian weight defined as in equation~\ref{eq: gaussian weight} with $\sigma$ a parameter of the influence of points in the neighborhood. $N_{x}$ can be approximated as a sphere of center $x$ and radius $3 \sigma$.
        
        The proposed UDF is simple. However, it was never previously tested as an input representation for neural scene completion. The intuitive approach to define an unsigned distance function would be to use the unsigned version of a classical SDF, such as IMLS~\cite{imls}, which we call UIMLS. IMLS is a weighted average of point-to-plane distances (see equation~\ref{eq: imls}), while UWED is a weighted average of Euclidean distances. As we show in the experiments section, UIMLS does not improve the learning of the scene completion network, meaning that general UDFs do not outperform signed ones.
        
        The proposed UDF is robust to noise and can be directly computed on a point cloud, without the need for normals estimation, which is especially effective when computed on noisy data, such as point clouds acquired using LiDAR sensors.
        
        
        After obtaining UWED computed on a 3D regular grid, we convert the values into voxel units and truncate the function at 3 voxels, following the work done in~\cite{scancomplete,sgnn}, obtaining the Truncated T-UWED on a sparse 3D grid.  
        
        Our experiments show that we obtain better geometry-learning ability by flipping the UWED function, as in~\cite{sscnet}: $\text{flipped T-UWED}(x) = 3 - \text{T-UWED}(x)$. We get smoother gradients between the T-UDF values and empty space in the sparse grid by overcoming the discontinuity of the function at truncation distance (visible in Figure~\ref{fig:TUDF_2D}).
    
    \subsection{Point cloud extraction from SDFs and UDFs} \label{sec: tudftopcl}
    
        Once the point cloud is transformed into an SDF or UDF on a sparse grid, we train the scene completion network and predict a new SDF or UDF representing the completed geometry of the scene. We first show how a point cloud can be extracted from an SDF, then we explain our method for extracting a point cloud from a UDF.
        
        Extracting an explicit geometry in the form of a point cloud makes it possible to compare the results directly with the original point cloud, unlike a mesh-to-mesh comparison, in which the ground truth mesh is constructed by meshing the original point cloud. Point clouds are also a good geometry proxy: it is an easy representation to exploit because no topology has to be considered. They can also be used directly for rendering with methods, such as splatting~\cite{splatting}, or more recently, Neural Point-Based Graphics~\cite{aliev2020}. 
        
        \subsubsection{Point cloud extraction from SDF}
        
            Marching Cubes~\cite{marching-cubes} is a method widely used for extracting a mesh from a signed distance field. It is also possible to use the same technique to extract a point cloud instead of a mesh. A point cloud can be obtained from an SDF computed on a regular grid, by linearly interpolating a 3D point using the SDF values on each edge between two consecutive voxels having opposite signs. The point cloud created corresponds exactly to the vertices of the mesh extracted using Marching Cubes~\cite{marching-cubes}. In our experiments, we use this method for point cloud extraction from the different SDFs.
        
        \subsubsection{Proposed point cloud extraction from UDF}
            We propose here a gradient approximation method to extract a point cloud from an unsigned distance field. We devise candidate selection criteria to check if an occupied voxel should be used to extract a point. For each occupied voxel $\mathbf{v}_{i,j,k}$ in the discretized grid, we check the UDF value of the current voxel $\text{udf}_{i,j,k}$ and the six neighboring voxels on the $x$, $y$, and $z$ axes. If all UDF values are defined (some voxels may not have UDF values because we are using sparse grids) and the UDF value of the current voxel is between $1 < \text{udf}_{i,j,k} < 3$ in voxel units, then it is selected as a candidate for extraction. The condition $\text{udf}_{i,j,k} > 1$ guarantees that all edges between the current voxel and its six neighboring voxels do not cross the surface.
            
            We then compute an approximation of the gradient of the unsigned distance field with finite differences and get the unit direction vector as follows: 
            \begin{equation}
                \mathbf{g}_{i,j,k} = 
                              \begin{bmatrix}
                                    \text{udf}_{i+1,j,k}-\text{udf}_{i-1,j,k} \\
                                    \text{udf}_{i,j+1,k}-\text{udf}_{i,j-1,k} \\
                                    \text{udf}_{i,j,k+1}-\text{udf}_{i,j,k-1}
                              \end{bmatrix}, \mathbf{d}_{i,j,k} = \frac{\mathbf{g}_{i,j,k}}{||\mathbf{g}_{i,j,k}||} 
            \end{equation} 
            
            We obtain the coordinates of the new extracted point on the surface by projection, through the use of the UDF value at the current voxel and the estimated direction vector:
            \begin{equation}
                \mathbf{p}_{new} = \mathbf{v}_{i,j,k} - \text{v\_{size}}  \; \text{udf}_{i,j,k} \; \mathbf{d}_{i,j,k} 
            \end{equation}
            where $\mathbf{v}_{i,j,k}$ is the 3D coordinates of the current voxel, $\text{v\_{size}}$ is the voxel size, and $\text{udf}_{i,j,k}$ is the scalar unsigned distance in voxel units.
            

\section{Experiments and Results}
    We claim that using the proposed UWED function as an input representation improves the scene completion task and helps neural networks to better understand the geometry of the scene. To validate our claim, we train two different networks on the task of scene completion. For the first network, we use SG-NN, which implements a U-Net architecture (taken from~\cite{sgnn}) with SparseConvNet convolutions (from~\cite{SubmanifoldSparseConvNet}) that regress distance fields on 3D grids. The second network, MinCompNet, is also a U-Net architecture, using Minkowski generalized convolutions~\cite{minkowski} with generative transposed convolutions~\cite{Gwak2020}. MinCompNet predicts only a binary occupancy grid. 
    
    As input to the networks, we use the different DF representations (signed and unsigned) and compare the predicted point clouds to target point clouds. For SG-NN, we extract the point cloud from the predicted DF representations with the methods explained in section~\ref{sec: tudftopcl}. For MinCompNet, we use the predicted occupied voxel coordinates as the predicted point cloud.

    We provide quantitative and qualitative results on the scene completion task, showing that our UDF representation performs well in the presence of noise, and large missing data. To this end, we experiment on a wide variety of point clouds, consisting of real indoor scenes acquired using RGB-D sensors (Matterport3D~\cite{Matterport3D}), synthetic outdoor scenes acquired with a simulated LiDAR using CARLA simulator~\cite{dosovitskiy2017} that we call PC3D-CARLA, and outdoor scenes acquired using a real mobile LiDAR system that we call PC3D-Paris. For the synthetic and real outdoor datasets, we use the Paris-CARLA-3D (PC3D) dataset~\cite{PC3D}. The different datasets tested have a wide variety of noise, missing data, and sparsity.

    \subsection{Distance Functions for Scene Completion}
        We compare eight DFs, which consist of signed and unsigned versions of point-to-plane and Euclidean distances.
        
        For SDFs, we compare Hoppe~\cite{Hoppe1992}, IMLS~\cite{imls}, Signed Weighted Euclidean Distance (SWED), which is the signed version of UWED, using IMLS for the sign, and Signed Euclidean Distance (SED), which is computed using 1 nearest neighboring point.
        
        For the UDFs, we test several functions in order to have a complete comparison with UWED. Specifically, we use the unsigned versions of the SDFs, obtaining UHoppe, UIMLS, UWED (the UDF introduced in section~\ref{sec: TUDF}), and Unsigned Euclidean Distance (UED), computed using 1 nearest neighboring point. 
        
        We convert all SDF and UDF values into voxel units and truncate them at 3 voxels. For notation simplification, we remove the T from the SDF and UDF names, but all tested DFs are truncated at 3 voxels and are defined on sparse 3D grids. First proposed in~\cite{sscnet}, a flipped version of T-SDF for completing partial scans, flips the function to remove strong gradients at truncation distance from the surface and give more signal to the network near the surface. Thus, we include the non-flipped and flipped versions of all the DFs in the comparisons.
        
        We do not test robust variants of SDFs such as FSS~\cite{fuhrmann2014} or RIMLS~\cite{oztireli2009} because they require more parameters (unlike all the functions tested, with at most one parameter $\sigma$), as the objective is to show the advantage of using unsigned distance fields as a weighted average of Euclidean distances over the most used SDFs like Hoppe or IMLS. 
    
    \subsection{Datasets Preparation}
        
        To effectively evaluate the different DFs on real data where we do not have ground truth, we follow the work done in SG-NN~\cite{sgnn}, in which a percentage of data is removed to create incomplete point clouds for the input, and all of the points are used for the target (see Figure~\ref{fig:scene_completion_pipeline}).
        
        To split the datasets, we use the original Matterport3D and PC3D splits train/val/test. 
        We remove 50\,\% of RGB-D frames and 90\,\% of LiDAR scans to build an incomplete input point cloud, compute the SDF or UDF values from it, and attempt to predict a new SDF or UDF using the DFs values computed on the original point cloud as the target. 
        
        For Matterport3D, we first convert the RGB-D frames into point clouds and keep the sensor positions to re-orient the computed normals needed for the SDFs. For training, we extract 12,000 point cloud blocks (64x64x128 grid size with 2\,cm voxel size) from training set rooms. The test set is composed of 6,000 point clouds from test set rooms.
        
        For PC3D-CARLA, we extract 3,000 blocks (128x128x128 grid size with 5\,cm voxel size) from each of the four training towns ($T_2$, $T_3$, $T_4$, $T_5$), resulting in 12,000 point cloud blocks for training. The test set is composed of 6,000 point clouds from towns $T_1$ and $T_7$.
        
        We follow the same procedure for PC3D-Paris, with 2,000 training point cloud blocks (128x128x128 grid size with 5\,cm voxel size) from $S_1$ and $S_2$. For PC3D-Paris, we use only the $S_3$ test set. As a result, the test set is composed of 1,000 point cloud blocks.
        
        We perform data augmentation, namely, random rotations around $z$, random scaling between 0.8 and 1.2, and local noise additions.
        
        We generate 4 hierarchical levels of the different distance fields to train the SG-NN network, by decreasing the resolution of the grid by a factor of two each time, and use only the highest resolution to train MinCompNet.
        
        For DFs that require normal computation on the point clouds (all except UED and UWED), we estimate the normal at each point using PCA with $n = 30$ neighbors and obtain a consistent orientation using point-wise RGB-D or LiDAR sensor positions.
        
    \subsection{Metrics for Scene Completion}
        Previous methods, such as ScanComplete~\cite{scancomplete} or SG-NN~\cite{sgnn} using SDFs for neural scene completion, compute the $\ell_1$ distance between the predicted SDF and the target SDF. However, $\ell_1$ results cannot be compared between different DFs.
        
        As we are extracting point clouds from the predicted SDFs and UDFs with SG-NN and from the predicted binary occupancy with MinCompNet, we can directly compare the extracted point cloud with the target cloud, which is the original raw point cloud.
        The most used metric for comparing point clouds is the Chamfer Distance (CD) between the predicted $P_1$ and original $P_2$ point clouds:
        \begin{equation}{\label{eq: CD}}
        \scriptsize
        CD = \frac{1}{2 |P_1|} \sum_{x \in P_1} \min_{y \in P_2} ||x - y||_{2} + \frac{1}{2 |P_2|} \sum_{y \in P_2} \min_{x \in P_1} ||y - x||_{2}
        \end{equation}
    
    \subsection{Implementation Details}
        We train SG-NN and MinCompNet using an NVIDIA GeForce RTX 2070 SUPER. For SG-NN, the loss is a combination of Binary Cross Entropy (BCE) on occupancy and $\ell_1$ loss on SDF or UDF predictions. For MinCompNet, the loss is the BCE on the occupancy grid. The trainings were done using ADAM optimizer, a learning rate of 0.001 and a batch size of 8 for both networks. We include the number of epochs and training time details in Table~\ref{tab: training epochs and time}. The training time is less for MinCompNet with a higher number of epochs than SG-NN due to the fact that the convolution operation is faster with MinkowskiEngine. The completion inference on GPU of a 128x128x128 grid (with a 5\,cm voxel size) and extraction of a point cloud on CPU is fast and takes around 0.6 seconds for SG-NN (resulting in point clouds with an average of 150,000 points) and 0.07 seconds for MinCompNet (resulting in point clouds with an average of 23,000 points).
        
        \begin{table}[ht!]
            \scriptsize
            \setlength{\tabcolsep}{1.5mm}
            \centering
            \begin{tabular}[t]{l|cc|cc}
                \centering
                 &  \multicolumn{2}{c|}{\textbf{SG-NN}} & \multicolumn{2}{c}{\textbf{MinCompNet}} \\
                \toprule
                \textbf{Datasets} & \textbf{\#epochs} & \textbf{Time} & \textbf{\#epochs} & \textbf{Time} \\
                \midrule
                Matterport3D& 5& 24h& 10& 9h \\
                \midrule
                PC3D-CARLA& 5& 24h& 10& 14h \\
                \midrule
                PC3D-Paris& 10& 10h& 50& 8h\\
                \bottomrule
            \end{tabular}
            \caption{Number of epochs and time taken to train each network on the different datasets. Time is reported in hours.}
            \label{tab: training epochs and time}
        \end{table}
    
    \subsection{Validity of the Proposed UDF as a Scene Representation} \label{sec: scene representation}
    
        Before comparing the influence of the different DFs on the scene completion networks, we evaluate the capacity of the proposed UDF to be a valid intermediate representation. To do so, we compute the error introduced by passing through the different DF representations and extracting a point cloud: point cloud$\,\to\,$DF$\,\to\,$point cloud. We then compute the CD between the original point cloud and the point clouds extracted from the different distance fields. We use the test sets of the three datasets with 6,000 point clouds for Matterport3D, 6,000 point clouds for PC3D-CARLA, and 1,000 point clouds for PC3D-Paris. Finally, we compute the mean CD over all point clouds.
        
\begin{figure}[ht!]
            \centering
                \includegraphics[width=\linewidth]{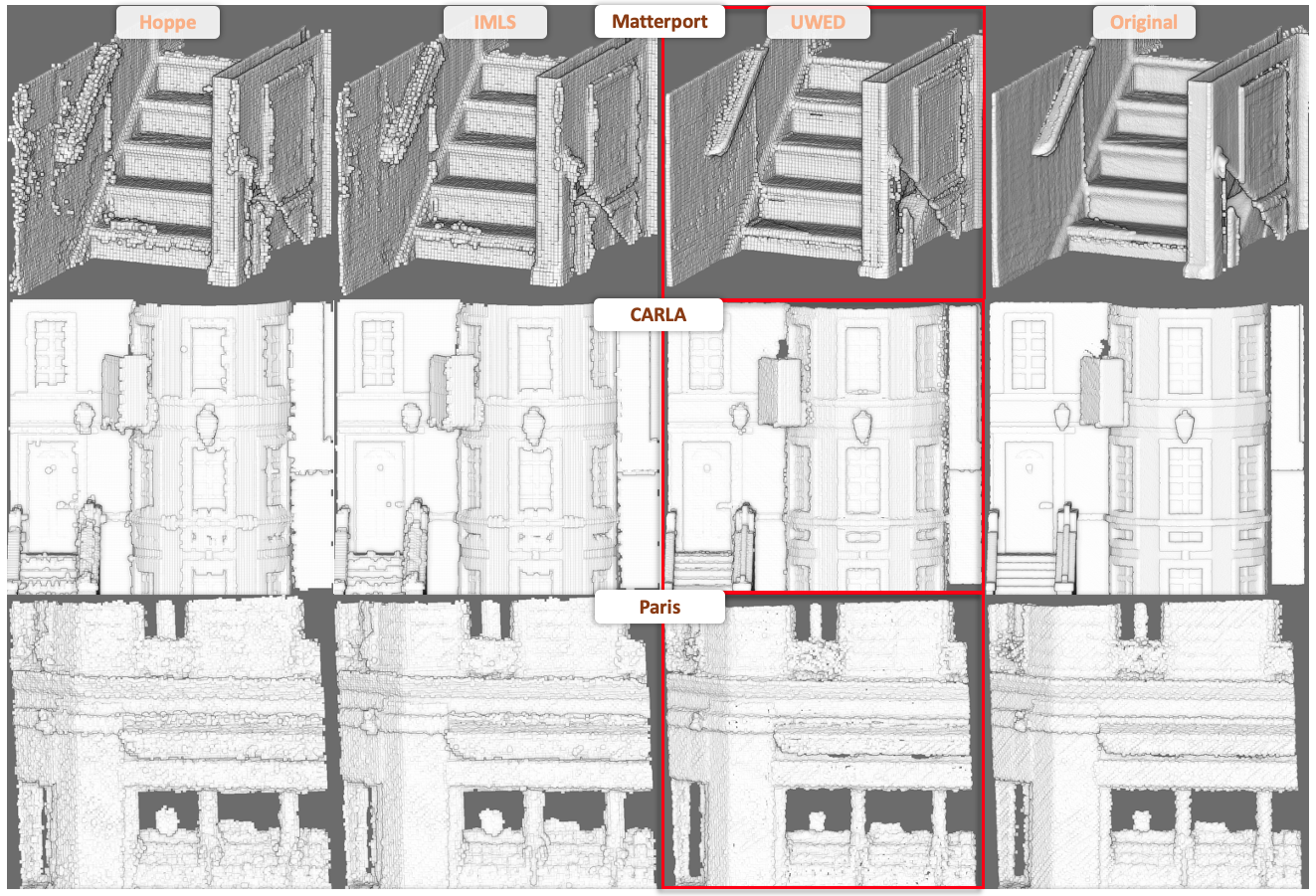}
                \caption{Comparison of different DF representations. From raw point clouds, we compute Signed Distance Functions (Hoppe and IMLS), and the proposed Unsigned Distance Function (UWED). Then we extract back a point cloud, without passing through a scene completion network, to compare the accuracy of the extracted point clouds to the original point cloud. With UWED, we are able to extract accurate point clouds, visible on the stair bar in Matterport and the window in the Paris dataset.}
                \label{fig: DFs-representation}
        \end{figure}

        Figure~\ref{fig: DFs-representation} shows the qualitative results of these comparisons. We can see that Hoppe is very noisy (only exploiting the closest point) and that Hoppe and IMLS expand the geometry at the borders. This is clearly visible on the stair bar for Matterport, the street lamp in the CARLA synthetic data, and the window jamb for the Paris data. On the contrary, with UWED, the extracted point cloud is very close to the original one.
       
       For quantitative evaluation, Table~\ref{tab: pcdfpc} shows the mean CD results on the three datasets. We can see that the lowest CD is obtained by UWED for Matterport3D and PC3D-Paris datasets and by UED for PC3D-CARLA dataset. PC3D-CARLA does not contain noise, it is a synthetic dataset acquired inside CARLA simulator, which is never the case in real-world datasets, such as PC3D-Paris and Matterport3D. The results also shows that our point cloud extraction method from UDFs, through approximate gradients by finite differences, allows the retrieval of an accurate point cloud.
        
        \begin{table}[!ht]
            \scriptsize
            \setlength{\tabcolsep}{0.6mm}
            \centering
            \begin{tabular}[t]{l|cccc|cccc}
                \centering
                 &  \multicolumn{4}{c|}{\textbf{SDFs}} & \multicolumn{4}{c}{\textbf{UDFs}} \\
                \toprule
                \textbf{Datasets} & \textbf{Hoppe} & \textbf{IMLS} & \textbf{SED} & \textbf{SWED} & \textbf{UHoppe} & \textbf{UIMLS} & \textbf{UED} & \textbf{UWED} \\
                \midrule
                Matterport3D & 0.90& 0.83& 0.93& \textbf{0.81}& 0.80&  0.74& 0.64& \textbf{0.60}\\
                \midrule
                PC3D-CARLA & 1.80& \textbf{1.78}& 1.90& 1.97& 1.67& 1.68& \textbf{1.52}& 1.62\\
                \midrule
                PC3D-Paris & 2.41& \textbf{2.25}& 2.38& 2.44& 2.90& 2.72& 2.24& \textbf{1.94}\\
                \bottomrule
            \end{tabular}
            \caption{Mean Chamfer Distance (in cm) between the extracted point clouds from the different DFs and the original point clouds, used to compare the accuracy of each function without passing through scene completion networks.}
            \label{tab: pcdfpc}
        \end{table}
        
        The parameter $\sigma$ used for IMLS, SWED, UIMLS, and UWED allows the functions to be robust to noise in the input data. A very small $\sigma$ makes the function sensitive to noise, while a large $\sigma$ results in smoothed sharp features. 
        
        To validate the choice of $\sigma$, in Table~\ref{tab: choice of sigma}, we provide quantitative results ranging $\sigma$ from one to four times voxel size comparing two DFs: IMLS (widely used SDF) and the proposed UWED function. We see the same trends between IMLS and UWED functions based on $\sigma$, showing that UWED will have the same behaviour with respect to the density of points as local distance functions like IMLS.
        
        For the following experiments, we fix $\sigma$ to two times the voxel size, the best trade-off for a fair comparison between the different DFs using $\sigma$.
        
        \begin{table}[ht!]
            \scriptsize
            \setlength{\tabcolsep}{1mm}
            \centering
            \begin{tabular}[t]{l|cccc}
                \centering
                 &  \multicolumn{4}{c}{\textbf{Variation of $\sigma$ for IMLS / UWED}} \\
                \toprule
                \textbf{Datasets} & \textbf{1 VS} & \textbf{2 VS} & \textbf{3 VS} & \textbf{4 VS} \\
                \midrule
                Matterport3D & 0.85/\textbf{0.60} & \textbf{0.83}/\textbf{0.60}& 0.87/0.61 & 0.87/0.61 \\
                \midrule
                PC3D-CARLA & \textbf{1.78}/1.96 & \textbf{1.78}/\textbf{1.62} & 1.82/1.75 & 1.88/1.83 \\
                \midrule
                PC3D-Paris & 2.32/2.42 & \textbf{2.25}/\textbf{1.94} & 2.68/2.08 & 2.60/2.23 \\
                \bottomrule
            \end{tabular}
            \caption{Mean Chamfer Distance (in cm) between raw and extracted point clouds from IMLS and UWED while varying the parameter $\sigma$ between one and four times the Voxel Size (VS).}
            \label{tab: choice of sigma}
        \end{table}
    
    \subsection{Results of Distance Functions on Scene Completion Networks}
    
        We compare the eight DFs and their flipped variants as input to the SG-NN network on the three datasets, which results in a total of 48 trainings. For MinCompNet network, we use three input representations for the comparisons, namely, occupancy grid, the flipped variants of IMLS, which is the most robust signed distance function, and UWED, resulting in 9 trainings. All quantitative results following training SG-NN are visible in Table~\ref{tab: quantitative-SG-NN}, while the results on MinCompNet are available in Table~\ref{tab: quantitative-MinCompNet}. We present the mean CD (equation~\ref{eq: CD}) across all test sets of the three datasets (6,000 for Matterport3D, 6,000 for PC3D-CARLA, and 1,000 point clouds for PC3D-Paris). 
        
        We can see in Table~\ref{tab: quantitative-SG-NN} that, for the SG-NN network, the flipped version of UWED results in a lower CD over all other distance functions for Matterport3D and PC3D-Paris datasets. We observe again that UED is better than UWED on PC3D-CARLA, which is expected, since PC3D-CARLA does not contain noise and the distance to the closest neighbor is more accurate that the weighted average of multiple neighboring points. However, UED is very sensitive to noise (Hoppe performs better than UED on noisy data such as Matterport3D and PC3D-Paris) and UWED is still better than any other SDF function on PC3D-CARLA. The unsigned version of IMLS (UIMLS) produces worse results than the signed version, meaning that in general, UDFs do not outperform SDFs. This stems from the fact that with UIMLS, the boundary expansion effect is still present (due to point-to-plane distances). This confirms the interest in using Euclidean distance in UWED formulation.
        
        We can note from Table~\ref{tab: quantitative-SG-NN} that systematically flipping the DFs improves results for UWED but not necessarily for SDFs. Indeed, for SDFs, the flipped function removes the discontinuity at the truncation distance and gets stronger gradients around the surface, however, it introduces a discontinuity at the surface level.

        \begin{table}[h!]
            \scriptsize
            \setlength{\tabcolsep}{1mm}
            \centering
            \begin{tabular}{l|c c|c c|c c}
            \toprule & \multicolumn{2}{c|}{\textbf{Matterport3D}} & \multicolumn{2}{c|}{\textbf{PC3D-CARLA} } & \multicolumn{2}{c}{\textbf{PC3D-Paris}} \\
                \midrule
                \midrule
                \multicolumn{7}{c}{\hskip0.5in \textbf{SDFs}} \\
                \midrule
                & \textsubscript{non-flipped} & \textsubscript{flipped} & \textsubscript{non-flipped} & \textsubscript{flipped} & \textsubscript{non-flipped} & \textsubscript{flipped} \\
                \midrule
                Hoppe & 1.54 & 1.53 & 6.32& 6.14 & 7.62 & 7.53  \\
                \midrule
                IMLS & 1.52 & \textbf{1.51} & 6.10 & 6.09 & \textbf{7.48} & 7.56  \\
                \midrule
                SED &  2.00&  2.01&  6.40&  6.84& 10.52&  10.34  \\
                \midrule
                SWED & 1.52 & 1.54 & \textbf{6.06} & 6.19 & 7.57 & 7.54 \\ \hline
                \midrule
                \multicolumn{7}{c}{\hskip0.5in \textbf{UDFs}} \\
                \midrule
                & \textsubscript{non-flipped} & \textsubscript{flipped} & \textsubscript{non-flipped} & \textsubscript{flipped} & \textsubscript{non-flipped} & \textsubscript{flipped} \\
                \midrule
                UHoppe & 1.25 & 1.26 & 15.24 & 6.00 & 9.95 & 7.96 \\
                \midrule
                UIMLS & 1.27 & 1.32 & 7.76 & 6.68 & 9.47 & 7.73 \\
                \midrule
                UED &  1.87&  1.80&  6.43&  \textbf{5.63}&  11.03& 10.11 \\
                \midrule
                UWED & 1.23 & \textbf{1.22} & 6.28 & 5.73 & 7.45 & \textbf{7.27} \\ 
                \bottomrule
            \end{tabular}
            \caption{SG-NN results for different SDFs and UDFs on three datasets. Results are the mean Chamfer Distance (in cm) computed on the test sets of the three datasets.}
            \label{tab: quantitative-SG-NN}
        \end{table}
        
        \begin{table}[ht!]
            \scriptsize
            \setlength{\tabcolsep}{1.5mm}
            \centering
            \begin{tabular}[t]{l|c|c|c}
                \centering
                 \diagbox{\textbf{Datasets}}{\textbf{Input}} & \multicolumn{1}{c|}{\textbf{Occupancy}} & \multicolumn{1}{c|}{\textbf{SDF IMLS}} & \multicolumn{1}{c}{\textbf{UDF UWED}} \\
                \toprule
                Matterport3D& 1.70& 1.75& \textbf{1.68}\\
                \midrule
                PC3D-CARLA& 5.67& 4.75& \textbf{3.98}\\
                \midrule
                PC3D-Paris& 4.67& 4.22& \textbf{4.0}\\
                \bottomrule
            \end{tabular}
            \caption{MinCompNet results using Binary Occupancy, IMLS, and UWED as input to predict occupancy grids, which are extracted as point clouds and compared to the original point clouds on three datasets. Results are the mean Chamfer Distance (in cm) computed on the test sets of the three datasets.}
            \label{tab: quantitative-MinCompNet}
        \end{table}
        
        \begin{figure}[h!]
            \centering
                \includegraphics[width=\linewidth]{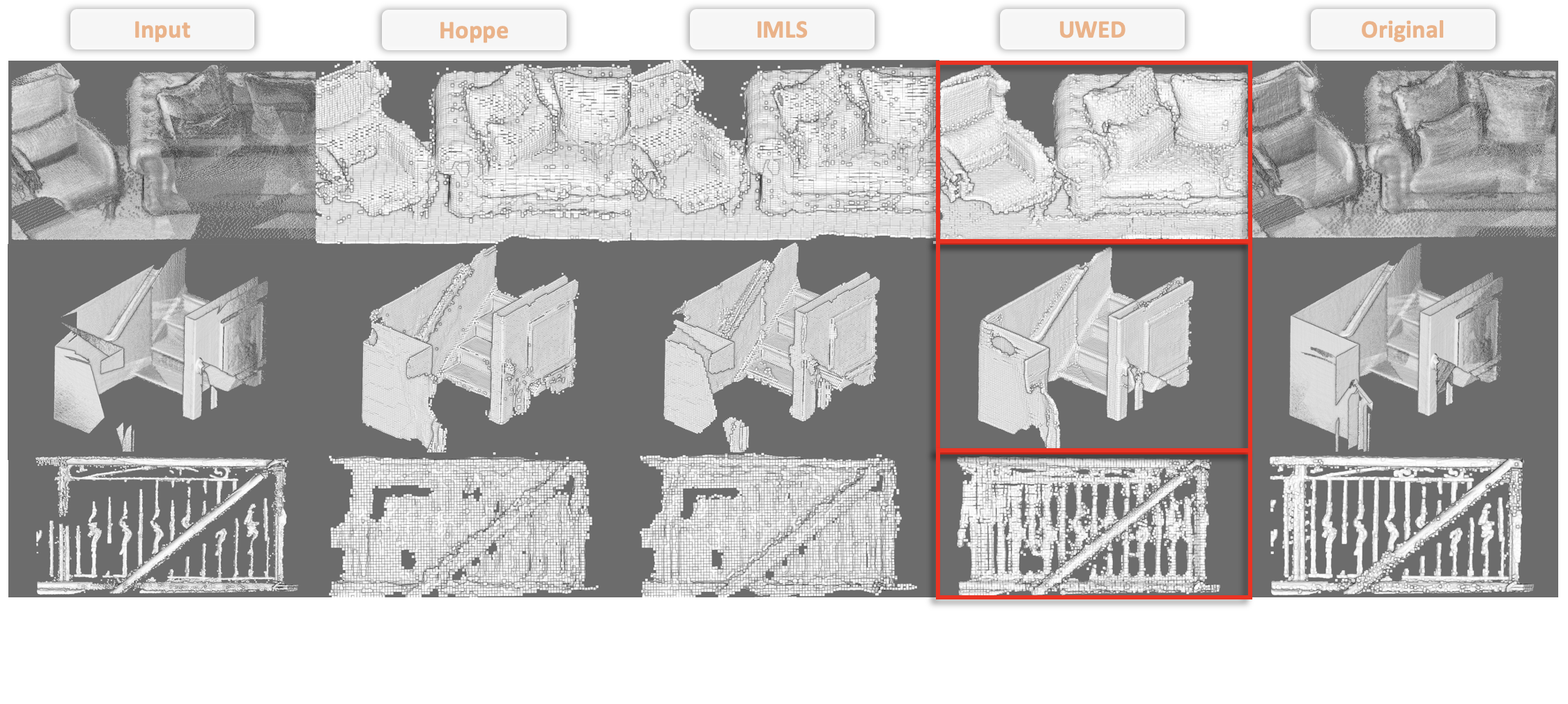}
                \caption{SG-NN inference using Hoppe, IMLS, and UWED on the Matterport3D test set.}
                \label{fig: SG-NN-predictions-matterport}
        \end{figure}
        
        \begin{figure}[h!]
            \centering
                \includegraphics[width=\linewidth]{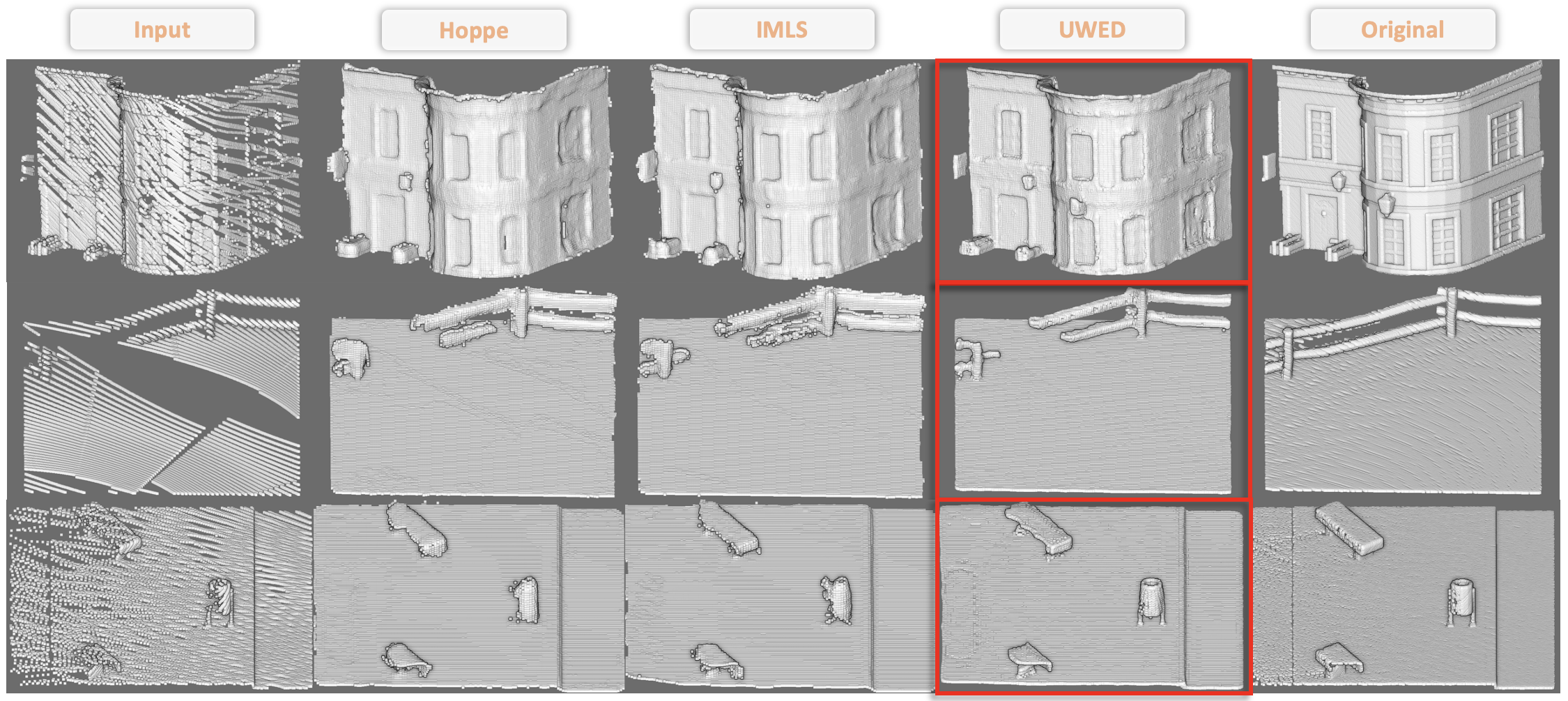}
                \caption{SG-NN inference using Hoppe, IMLS, and UWED on the PC3D-CARLA test set.}
                \label{fig: SG-NN-predictions-carla}
        \end{figure}
        
        \begin{figure}[h!]
            \centering
                \includegraphics[width=\linewidth]{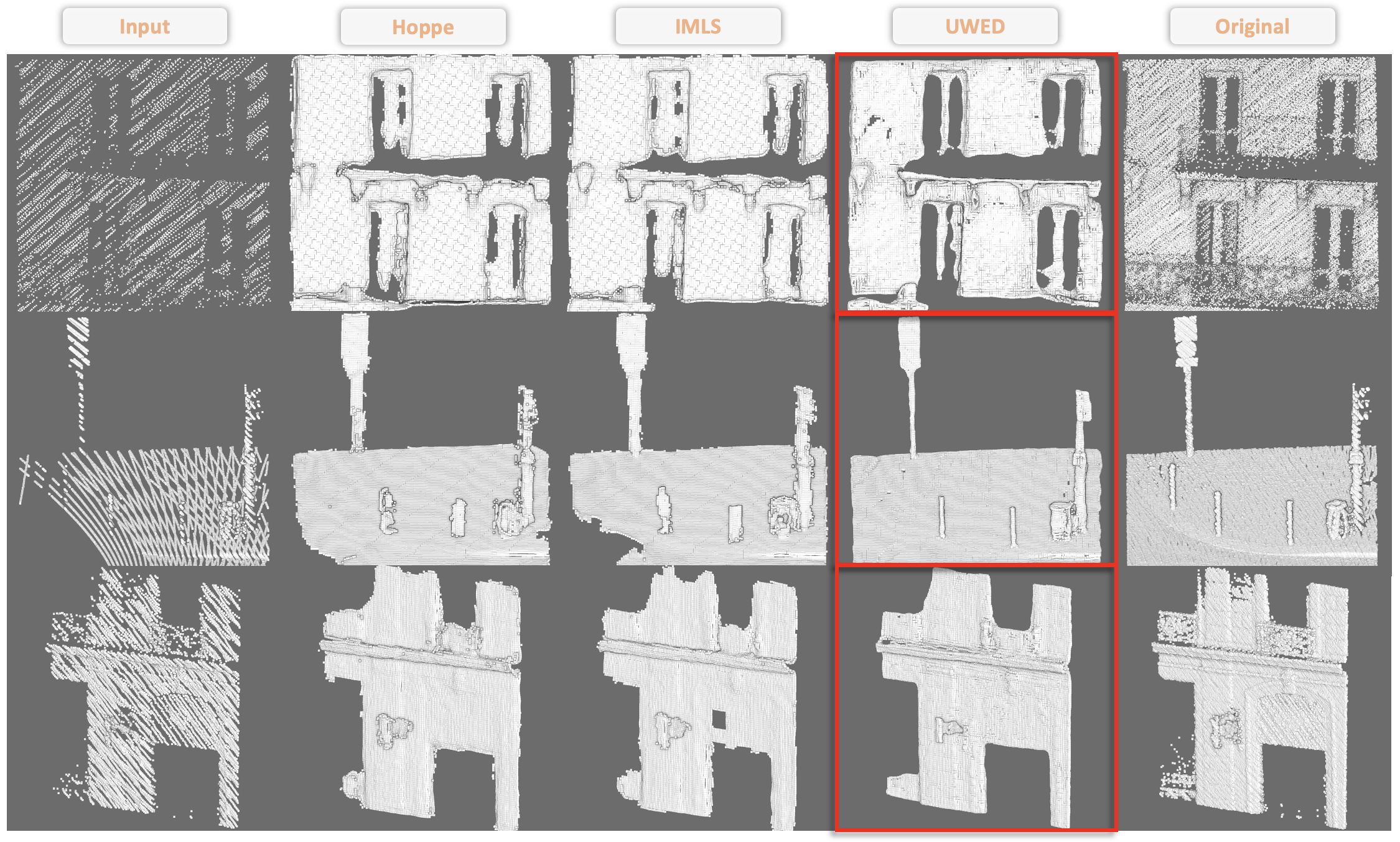}
                \caption{SG-NN inference using Hoppe, IMLS, and UWED on the PC3D-Paris test set.}
                \label{fig: SG-NN-predictions-soufflot}
        \end{figure}
        
\begin{figure}[h]
            \centering
                \includegraphics[width=\linewidth]{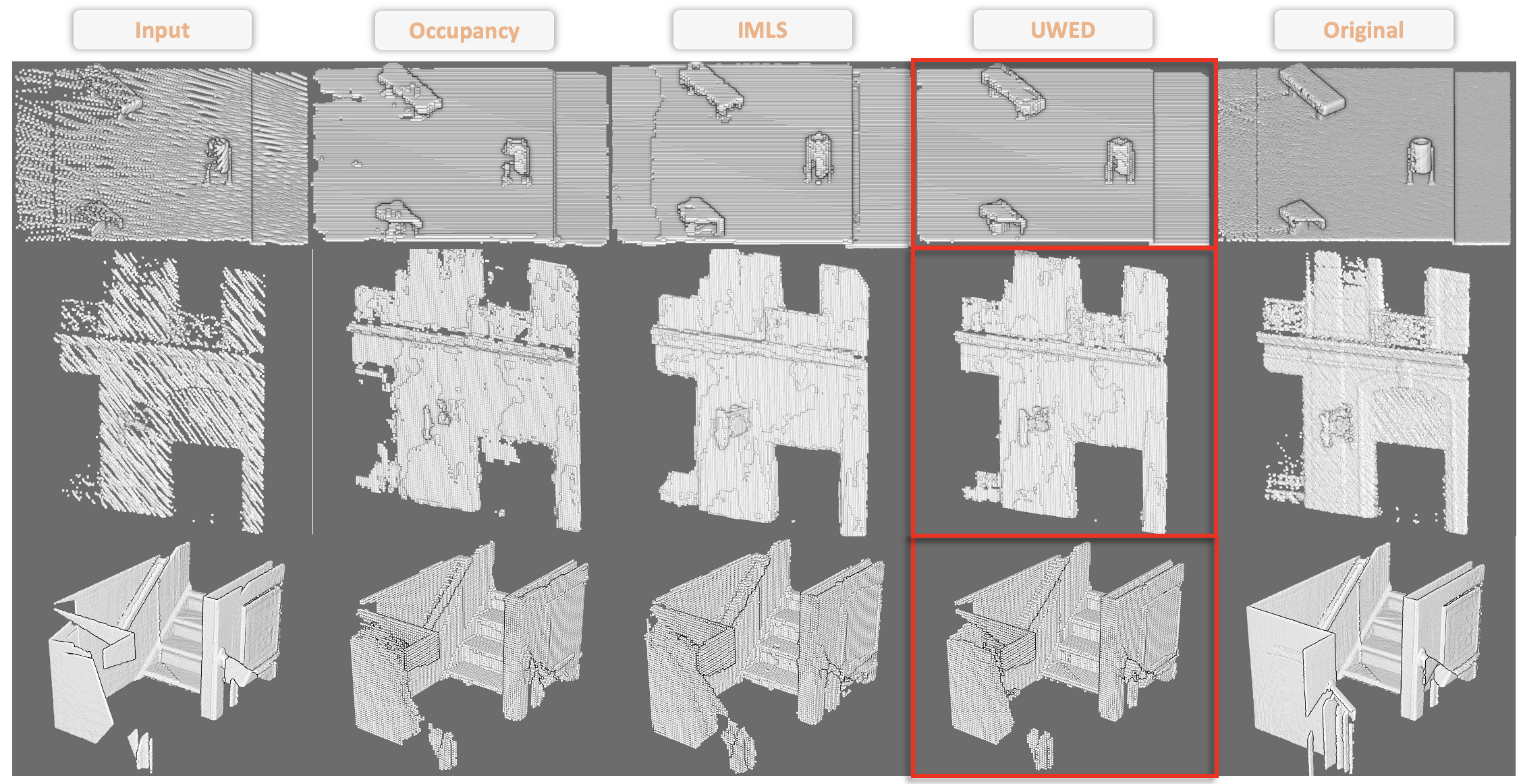}
                \caption{MinCompNet inference using Occupancy, IMLS, and UWED as input, and predicting occupancy on PC3D-CARLA, PC3D-Paris, and Matterport3D test sets from top to bottom.}
                \label{fig: MinCompNet-predictions}
        \end{figure}

        Figures~\ref{fig: SG-NN-predictions-matterport},~\ref{fig: SG-NN-predictions-carla}, and~\ref{fig: SG-NN-predictions-soufflot} show qualitative results of SG-NN inferences for Hoppe (well-used SDF), IMLS (best SDF in Table~\ref{tab: quantitative-SG-NN}), and UWED. Figure~\ref{fig: MinCompNet-predictions} shows qualitative results of MinCompNet inferences for occupancy, IMLS, and UWED. All DFs used for qualitative evaluations are flipped versions. We observe each time that higher quality point clouds are obtained after scene completion using the proposed UDF representation, which is visible on stairs, trash can, and bollards. UWED is also able to complete larger missing regions (middle example, Figure~\ref{fig: SG-NN-predictions-soufflot}). Comparing the results in Tables~\ref{tab: quantitative-SG-NN} and~\ref{tab: quantitative-MinCompNet}, we notice that better scene completion is achieved with SG-NN on Matteport3D, since the network was designed to perform completion on the same indoor RGB-D dataset, while MinCompNet has a lower CD on the two other datasets (made of LiDAR scans with sparser data). MinCompNet seems to be able to complete larger missing areas thanks to the generative transposed convolutions~\cite{Gwak2020}, however, SG-NN still has the advantage of more accurate geometry prediction (see UWED in third row of Figure~\ref{fig: SG-NN-predictions-carla} and first row of Figure~\ref{fig: MinCompNet-predictions}), due to the fact that it predicts a continuous DF representation. In all cases, the proposed input representation gives better results than a binary occupancy grid or a classical SDF.
        
    \subsection{Limitations and Perspectives}
        The main limitation of the proposed representation is that it does not allow the direct extraction of a mesh, unlike SDFs. The purpose of UWED is not to reconstruct the surface, but to get an intermediate representation that gives better signal to scene completion networks. 
        
        In perspective, UWED demonstrated a high performance on the scene completion task. Following the results, UWED seems to allow neural networks to better learn the geometry. Thus, it should be possible to use it for other tasks in 3D (registration, semantic segmentation, object detection). However, the SOTA networks on theses tasks are designed to work on points clouds and not on SDF, like scene completion networks. Using UWED for these tasks might require some architecture changes, which we leave for future work.

\section{Conclusion}
    We presented an Unsigned Distance Function to be used as input to neural scene completion networks that improves the quality of the completion task. We also introduced a point cloud extraction algorithm from an UDF computed on a sparse grid. Finally, we compared eight different SDFs and UDFs on three different datasets. Our experiments showed quantitatively and qualitatively that UWED can be used as a neural scene completion input representation and achieves the best completion results with two state-of-the-art scene completion networks.

{\small
\bibliographystyle{ieee_fullname}
\bibliography{UWED}
}

\end{document}